\documentclass[runningheads]{llncs}
\usepackage{graphicx}
\usepackage{comment}
\usepackage{threeparttable}
\usepackage{amsmath,amssymb} 
\usepackage{color}
\usepackage{hyperref}

\hypersetup{
    colorlinks=true,
    linkcolor=red,
    filecolor=magenta,      
    urlcolor=blue,
}

\begin{document}
\pagestyle{headings}
\mainmatter
\def\ECCVSubNumber{1972}  

\title{Adaptive Offline Quintuplet Loss for Image-Text Matching} 

\titlerunning{Adaptive Offline Quintuplet Loss}
%
\author{Tianlang Chen\inst{1}\orcidID{0000-0002-6355-6474} \and
Jiajun Deng\inst{2} \and
Jiebo Luo\inst{1}}
\authorrunning{Tianlang Chen, Jiajun Deng, Jiebo Luo}
%
\institute{University of Rochester,\email{ \{tchen45,jluo\}@cs.rochester.edu}, \and
    Uiversity of Science and Technology of China,\email{ \{djiajun1206\}@gmail.com}}
\maketitle

\begin{abstract}
Existing image-text matching approaches typically leverage triplet loss with online hard negatives to train the model. For each image or text anchor in a training mini-batch, the model is trained to distinguish between a positive and the most confusing negative of the anchor mined from the mini-batch (\emph{i.e.} online hard negative). This strategy improves the model's capacity to discover fine-grained correspondences and non-correspondences between image and text inputs. However, the above approach has the following drawbacks: (1) the negative selection strategy still provides limited chances for the model to learn from very hard-to-distinguish cases. (2) The trained model has weak generalization capability from the training set to the testing set. (3) The penalty lacks hierarchy and adaptiveness for hard negatives with different ``hardness'' degrees. In this paper, we propose solutions by sampling negatives offline from the whole training set. It provides ``harder'' offline negatives than online hard negatives for the model to distinguish. Based on the offline hard negatives, a quintuplet loss is proposed to improve the model's generalization capability to distinguish positives and negatives. In addition, a novel loss function that combines the knowledge of positives, offline hard negatives and online hard negatives is created. It leverages offline hard negatives as the intermediary to adaptively penalize them based on their distance relations to the anchor. We evaluate the proposed training approach on three state-of-the-art image-text models on the MS-COCO and Flickr30K datasets. Significant performance improvements are observed for all the models, 
proving the effectiveness and generality of our approach. Code is available at \url{https://github.com/sunnychencool/AOQ}.

\end{abstract}

\section{Introduction} \label{sec:intro}

Image-text matching is the core task in cross-modality retrieval to measure the similarity score between an image and a text. By image-text matching,
a system can retrieve the top corresponding images of a sentence query, or retrieve the top corresponding sentences of an image query.

To train an image-text matching model to predict accurate similarity score, triplet loss is widely used \cite{nam2017dual,eisenschtat2017linking,faghri2017vse++,li2019visual,lee2018stacked}. Each given image or text of a training mini-batch is referred to as an \textit{anchor}. For each image/text anchor, a text/image that corresponds to the anchor is called a \textit{positive} while one that does not correspond to the anchor is called a \textit{negative}. The anchor and its positives/negatives belong to two modalities. A triplet loss is applied to encourage the model to predict higher similarity scores between the anchor and its positives (\emph{i.e.} positive pairs) than those between the anchor and its negatives (\emph{i.e.} negative pairs).

To utilize negative pairs to train the model, early approaches \cite{nam2017dual,eisenschtat2017linking,huang2017instance} adopt an all-in strategy. For each anchor, all its negatives in the mini-batch participate in the loss computing process. However, in most situations, the semantic meanings of an anchor and its negatives are totally different. With this strategy, the overall training difficulty is relatively low for the model to distinguish between positive and negative pairs. The model only needs to focus on each pair's global semantic meaning difference and may ignore the local matching details. Faghri et al. \cite{faghri2017vse++} propose a triplet loss with online hard negatives (\emph{i.e.} online triplet loss) as a more effective training approach. Specifically, for each anchor in a mini-batch, the model computes its similarity score to all the negatives in the same mini-batch online, and selects the negative with the highest score to the anchor as online hard negative of the anchor. The new triplet loss guides the model to only distinguish between the positives and online hard negatives of the anchor. Compared with the all-in strategy, the models trained by this approach commonly achieve better performance in distinguishing between positives and confusing negatives that have similar semantic meanings to the anchor. This training approach is employed by all the state-of-the-art models \cite{li2019visual,lee2018stacked,liu2019focus,wang2019camp}. 

Even with its effectiveness, we argue that the online triplet loss still have three drawbacks in negative selection strategy, distinguishing strategy, and penalization strategy: (1) for the negative selection strategy, the ``hardness'' degree of online hard negatives is still not sufficient. Given the MS-COCO dataset as example, the training set contains 500K corresponding image-text pairs. When we set the mini-batch size to 128 as in \cite{li2019visual,lee2018stacked,liu2019focus,wang2019camp}, for each online hard negative of an anchor mined from the mini-batch, we can prove that its similarity score rank expectation to the anchor in the whole training set is about $4000$ (\emph{i.e.} $\frac{500K}{128}$). The probability of its rank in the  top-$100$ is only about 2.2\%. In other words, a very hard negative with a top-100 similarity score rank for the anchor will rarely be sampled to train the model. This decreases the model's capacity to distinguish between the positives and those very confusing negatives. Increasing the mini-batch size could be helpful. However, the mini-batch computational complexity grows sharply. (2) For the distinguishing strategy, the triplet loss only focuses on obtaining the correct rank orders between the positives and negatives of the same anchor. However, it does not guide the model to rank among positive pairs and negative pairs that contain no common samples. Actually, this guidance is essential to improve the model's generalization capability from training to testing, especially when we apply the guidance on the very hard negative pairs. (3) For the penalization strategy, the triplet loss lacks a hierarchy. Ideally, the loss function should guide the model to maintain remarkable score gaps among the pairs of different classes. For example, the positive pairs should obtain far higher similarity scores than very hard negative pairs, and the very hard negative pairs should also obtain far higher similarity scores than ordinary hard negative pairs. When a pair's predicted score is close or beyond the boundary of its pair class, the loss function should give it a larger penalty to update the model. However, the current online triplet loss only defines positive and online hard negative pairs. More importantly, it gives an equal penalty to all the pairs when the margin conditions are not satisfied. 

\begin{figure}[!t]

\centering
\includegraphics[width=0.79\columnwidth]{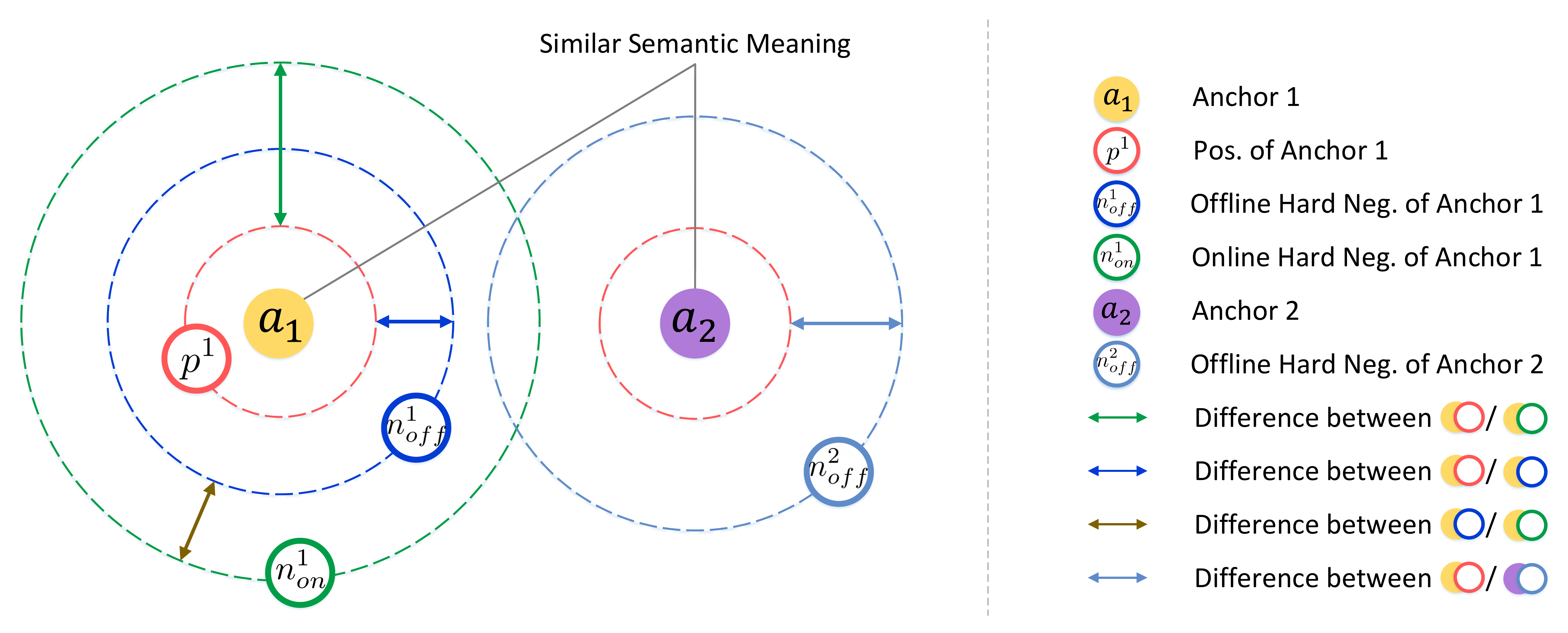}

\caption{Overview of the proposed training approach. For each anchor, we sample its positives, offline hard negatives and online hard negatives. The training approach gives adaptive penalties to enlarge the similarity score differences among positive pairs, offline hard negative pairs and online hard negative pairs (\emph{i.e.} the blue, green and brown arrows). On the other hand, extra penalties are added to enlarge the similarity score difference between positive pairs and offline hard negative pairs with different anchors that share similar semantic meanings (\emph{i.e.} the cyan arrow). }
\label{fig:overview}

\end{figure}

To overcome the above drawbacks, we propose a new training approach that can be generally applied on {\it all} existing models. Specifically, we utilize a two-round training to additionally sample ``harder'' negatives offline. In the first round, we train the model by the original online triplet loss. After that, for each image and text anchor in the training set, the model predicts its similarity score to all its negatives in the training set and ranks them. In the second round, given each anchor in a mini-batch, we sample its offline hard negatives directly from its top negative list with the highest similarity score in the whole training set. In this process, multiple kinds of offline hard negative pairs are constructed which share/do not share common elements with the positive pairs. The model is trained by a combination of online triplet loss and offline quintuplet loss to overcome the first two drawbacks successfully. Furthermore, we modify the loss function and feed information of offline hard negative pairs into the online triplet loss term. The complete training loss achieves hierarchical and adaptive penalization for the positive pairs, offline hard negative pairs, and online hard negative pairs with different ``hardness'' degrees. The framework of the proposed training approach is shown in Figure~\ref{fig:overview}.

Our main contributions are summarized as follows:

\begin{itemize}
    \item We propose a novel and general training approach for image-text matching models. A new offline quintuplet loss is introduced that can effectively cooperate with the original online triplet loss.

    \item We skillfully feed the similarity score of offline hard negative pair into online loss term. It serves as a criterion to adaptively penalize different kinds of pairs. We analyze how it works mathematically.

    \item  We evaluate our training approach on three state-of-the-art image-text matching models. Quantitative and qualitative experiments conducted on two publicly available datasets demonstrate its strong generality and effectiveness. 
\end{itemize}

\section{Related Work}

Image-text matching has received much attention in recent years. Most of the previous works focus on the improvement of feature extraction and model design. Early image-text matching approaches \cite{frome2013devise,kiros2014unifying,faghri2017vse++,zhang2018deep} directly capture the visual-textual alignment at the level of image and text. Typically, they extract the global image feature by convolutional neural network (CNN), and extract the global text feature by language model such as Skip-gram model \cite{mikolov2013efficient} or recurrent neural network (RNN). The image-text similarity score is then computed as the inner product \cite{frome2013devise,kiros2014unifying,faghri2017vse++} or cosine similarity \cite{zhang2018deep} of the image and text features. The success of attention models for joint visual-textual learning tasks, such as visual question answering (VQA) \cite{yu2017multi,lu2016hierarchical,yang2016stacked,kim2016multimodal} and image captioning \cite{xu2015show,lu2017knowing,you2016image,pedersoli2017areas,chen2018factual}, leads to the transition to capture image-text correspondence at the level of image regions and words \cite{huang2017instance,li2017identity,nam2017dual,zheng2017dual}. Typically, these approaches extract the image region feature and word feature from the last pooling layer of CNN and temporal outputs of RNN. They focus on designing effective upper networks that can automatically find, align and aggregate corresponding regions and words to compute the final similarity score. Recently, Anderson et al. \cite{anderson2018bottom} extract the image object features by the combination of Faster R-CNN \cite{ren2015faster} and ResNet \cite{he2016deep} for VQA. Based on \cite{anderson2018bottom}, recent approaches \cite{lee2018stacked,li2019visual,liu2019focus,wang2019camp,ji2019saliency} further construct the connection between words and image objects. They either propose new mechanisms for object feature extraction, such as feeding saliency information \cite{ji2019saliency} or extracting joint features among objects by constructing object graph \cite{li2019visual}, or propose different cross-modality aggregation networks \cite{lee2018stacked,wang2019camp,liu2019focus,chen2020expressing,huang2019acmm} to improve the aggregation process from object and word features to the final score.   

Even though the network design is widely studied, relatively fewer works focus on the training approach. Early image-text matching approaches \cite{frome2013devise,kiros2014unifying,eisenschtat2017linking,you2018end} commonly apply a standard triplet loss whose early form can be found in \cite{weston2010large} for word-image embedding. On the other hand, Zhang et al. \cite{zhang2018deep} improve the triplet loss and propose a norm-softmax loss to achieve cross-modal projection. For both losses, all the negatives of an anchor in the same mini-batch are utilized for loss computing. Significant improvement is observed as Faghri et al. \cite{faghri2017vse++} propose the triplet loss with online hard negatives. Online triplet mining is first introduced in \cite{schroff2015facenet} for face recognition. For image-text matching, it mines the online hard negatives of the anchors from the mini-batch and makes the model only pay attention to these confusing negatives. Almost all the current models \cite{li2019visual,lee2018stacked,liu2019focus,wang2019camp} apply this online triplet loss. To the best of our knowledge, our work is the first that introduces offline hard negatives for image-text matching. They are mined offline from the whole training set. Motivated by \cite{chen2017beyond} for person re-identification, we propose a quintuplet loss based on offline hard negatives to effectively cooperate with an online triplet loss, leading to significant improvement. It should be noticed that Liu et al. \cite{liu2019hal} explicitly feed adaptive penalty weight into triplet loss for image-text matching. However, they use it to solve the hubness problem, while we implicitly feed hierarchical information into the model to enlarge the similarity score differences among different pair classes.

\section{Methods}

\begin{figure}[!t]

\centering
\includegraphics[width=0.99\columnwidth]{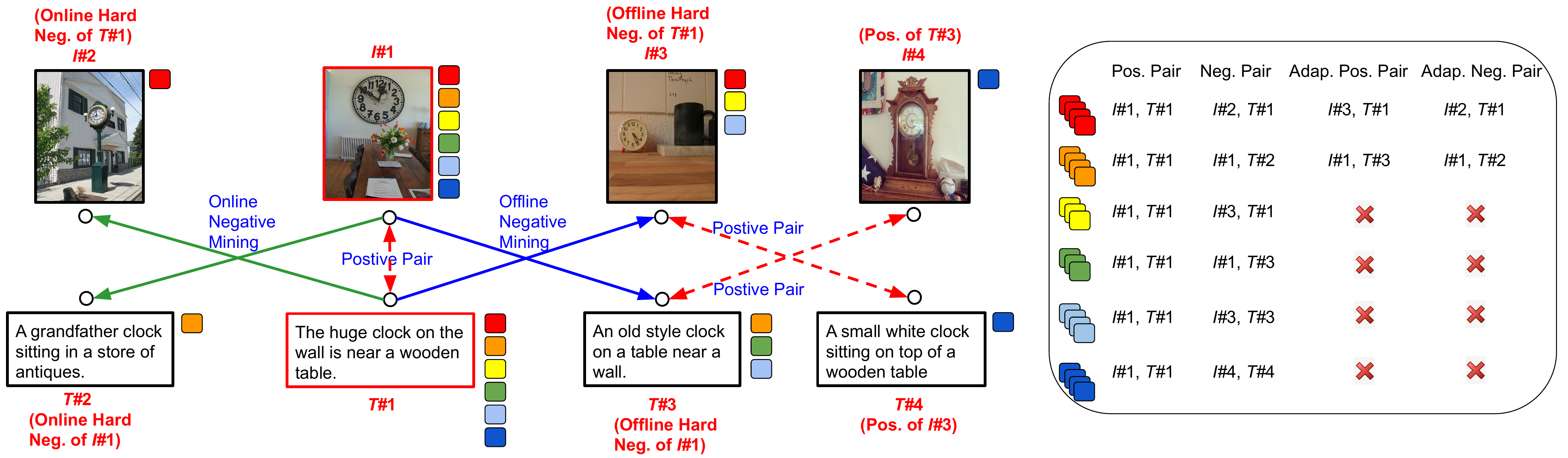}

\caption{Training process illustration. Given a positive image-text pair $(I\#1, T\#1)$, 6 margin-based ranking losses are applied to enlarge its similarity score differences from the online hard negative pairs $(I\#2, T\#1)$, $(I\#1, T\#2)$, the offline hard negative pairs $(I\#3, T\#1)$, $(I\#1, T\#3)$ (with the common anchor), and the derived offline hard negative pairs $(I\#3, T\#3)$, $(I\#4, T\#4)$ (without the common anchor). Adaptive penalization is imposed via the online losses to adaptively penalize positive and negative pairs with different strengths and directions. The involved samples of each loss are marked by the corresponding squares.}
\label{fig:model}

\end{figure}

In this section, we formally present our training approach for image-text matching. In Section~\ref{sec:basic}, we introduce the margin-based standard and online triplet losses that are used in previous works. In Section~\ref{sec:offline}, we present offline quintuplet loss as an effective complement to online triplet loss to significantly improve the performance. In Section~\ref{sec:adap}, we propose our final loss function with adaptive penalization and mathematically show how it works. The overall training process and the involved pairs are illustrated in Figure~\ref{fig:model}.

\subsection{Triplet Loss for Image-text Matching} \label{sec:basic}

Given an input image-text pair, image-text matching models aim to predict the pair's similarity score as a criterion for cross-modality retrieval. To achieve this, positive pairs (\emph{i.e.} corresponding image-text pairs) and negative pairs (\emph{i.e.} non-corresponding image-text pairs) are constructed. The model is trained to predict higher similarity score for the positive pairs than the negative ones.

Because the metrics of cross-modality retrieval are based on the ranking performance of multiple candidates on a single query, triplet loss is widely applied to train the model. It holds a common sample for each positive pair and negative pair as an \textit{anchor}. The other sample in the positive pair is called the anchor's \textit{positive} while the other sample in the negative pair is called the anchor's \textit{negative}. In essence, triplet loss encourages the model to predict higher similarity scores from the anchor to its positives. This is consistent with the retrieval process of finding the corresponding candidates of a query with the high similarity scores.

Early image-text matching works \cite{frome2013devise,kiros2014unifying,eisenschtat2017linking,you2018end} typically apply a standard triplet loss without hard negative mining. Given a training mini-batch that contains a set of positive pairs, the standard triplet loss is defined as:

\begin{equation} \label{equ:stdloss}
\begin{aligned}
\mathcal{L}_{std} = \sum_{(i,t) \in P}(\sum_{\overline{t} \in T/{t}}[\gamma - S(i,t)+S(i,\overline{t})]_{+}  + 
\sum_{\overline{i} \in I/{i}}[\gamma - S(i,t)+S(\overline{i},t)]_{+})
\end{aligned}
\end{equation}
Here $\gamma$ is the margin of the triplet loss, $[x]_{+} \equiv max(x, 0)$. $I$, $T$ and $P$ are the image, text and positive pair sets of the mini-batch, respectively. $i$ and $t$ are the anchors of the two terms, respectively. $(i, t)$ represents the positive pair, while  $(i, \overline{t})$ and $(\overline{i},t)$ represent the negative pairs available in the mini-batch.

On the other hand, to overcome the drawback of standard triplet loss mentioned in Section~\ref{sec:intro}, Faghri et al. \cite{faghri2017vse++} present triplet loss with online hard negatives (\emph{i.e.} online triplet loss). In particular, for a positive pair $(i, t)$ in a mini-batch, the hard negatives of the anchor $i$ and $t$ are given by $\overline{t}_{on} = argmax_{c \in T/{t}}S(i,c)$ and $\overline{i}_{on} = argmax_{b \in I/{i}}S(b,t)$, respectively. The online triplet loss is defined as:

\begin{equation} \label{equ:online}
\begin{aligned}
\mathcal{L}_{online} = \sum_{(i,t) \in P}([\gamma - S(i,t)+S(i,\overline{t}_{on})]_{+}  + [\gamma - S(i,t)+S(\overline{i}_{on},t)]_{+}) 
\end{aligned}
\end{equation}
Compared with the standard triplet loss, online triplet loss forces the model to only learn to distinguish between the positive and the most confusing negative of an anchor in the mini-batch. This guides the model to not only consider the overall semantic meaning difference of a pair, but also discover correspondences and non-correspondences from the details hidden in local regions and words.

\subsection{Offline Quintuplet Loss} \label{sec:offline}

\begin{figure}[!t]

\centering
\includegraphics[width=0.95\columnwidth]{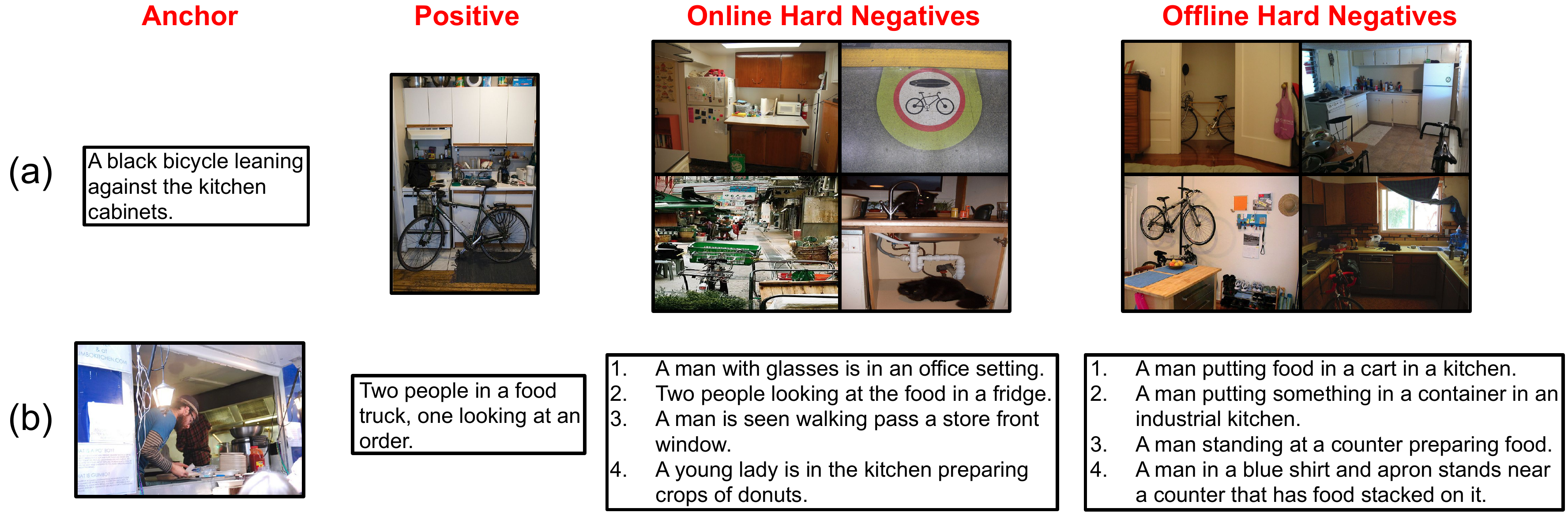}

\caption{Two example anchors, their corresponding positives, their sampled online hard negatives and offline hard negatives.}
\label{fig:negatives}

\end{figure}

One problem of online triplet loss in Section~\ref{sec:basic} is that the ``hardness'' degree of most online hard negatives is still not sufficient, especially when the training involves a large-scale training set and a relatively small batch size. As mentioned in Section~\ref{sec:intro}, the rank of an anchor's online hard negative in the whole training set is commonly not very high. Qualitatively, as shown in Figure~\ref{fig:negatives}, the online hard negatives of an anchor typically contain a few related words, objects or scenes to the anchor. However, there exist obvious non-correspondences between the anchor and the negatives. Indeed, the model only needs to find these non-correspondences and strengthen their influence, which is sufficient for the score difference between the positive pair and negative pair to exceed the margin $\gamma$ in Equation~\ref{equ:online}. However, during inference, when the model encounters ``harder'' negatives like the offline hard negative examples of Figure~\ref{fig:negatives}, the model may not be able to distinguish them from the positives. The non-corresponding parts of these ``harder'' negatives to the anchor are subtle, and their influence on the predicted score can be offset by the perfectly corresponding parts.

To overcome the problem, we additionally mine ``harder'' negatives in an offline fashion. In particular, it involves a two-round training. In the first round, the model is trained by the online triplet loss. After that, it performs global similarity score prediction -- for each image/text in the training set, the model predicts its similarity score to all its non-corresponding texts/images in the training set, ranks them by their scores and stores the list of the top-$h$. In the second round, for each anchor in a mini-batch, its offline hard negatives are uniformly sampled from the top-$h$ negatives of the anchor in the whole training set. The model is trained from scratch again by the following loss function:

\begin{equation} \label{equ:onoffbasic}
\scalebox{0.92}{
$\begin{aligned}
\mathcal{L} = \sum_{(i,t) \in P}(([\gamma_{1} - S(i,t)+S(i,\overline{t}_{on})]_{+} + [\gamma_{2} - S(i,t)+S(i,\overline{t}_{off})]_{+})  \\ + 
([\gamma_{1} - S(i,t)+S(\overline{i}_{on},t)]_{+} + [\gamma_{2} - S(i,t)+S(\overline{i}_{off},t)]_{+}))
\end{aligned}$}
\end{equation}
Here $\overline{t}_{off}$ and $\overline{i}_{off}$ are the offline hard negatives of $i$ and $t$, $\gamma_{1}$ and $\gamma_{2}$ are the margins of the online and offline triplet losses. It should be noticed that for models with relatively low inference speed, the above mentioned global similarity score prediction step can be time-consuming. In Section~\ref{sec:exp}, we demonstrate that a model can safely utilize the prediction of another efficient model to mine offline hard negatives, which still sharply benefits the training process.    

Because the offline hard negatives are very confusing, to make them benefit the training, we should set $\gamma_{2}$ to a lower margin than $\gamma_{1}$, \emph{e.g.} $0$. However, in this situation, if the positive and offline hard negative pairs share a same anchor, the model will merely learn how to find the subtle non-corresponding parts of the offline hard negative pair, but still does not learn how to deal with the situation when the negative pair's perfect matching parts offset the score influence of non-corresponding parts. We attribute it to the fact that the positive and offline hard negative get close similarity score for their corresponding parts to the same anchor. The model only needs to find the non-corresponding parts of the negative pair to satisfy the margin condition of $\gamma_{2}$. Also, as claimed in \cite{chen2017beyond}, this setting weakens the model's generalization capability from training to testing.

Considering this, we additionally derive two offline hard negative pairs and modify Equation~\ref{equ:onoffbasic} for the second-round training as follows:
\begin{equation} \label{equ:onoff}
\scalebox{0.75}{
$\begin{aligned}
\mathcal{L}= \sum_{(i,t) \in P}(([\gamma_{1} - S(i,t)+S(i,\overline{t}_{on})]_{+} + [\gamma_{2} - S(i,t)+S(i,\overline{t}_{off})]_{+} + [\gamma_{2} - S(i,t)+S(\overline{i}_{off},\overline{t}_{off})]_{+})  \\ + 
([\gamma_{1} - S(i,t)+S(\overline{i}_{on},t)]_{+} + [\gamma_{2} - S(i,t)+S(\overline{i}_{off},t)]_{+} + [\gamma_{2} - S(i,t)+S(\widetilde{\overline{i}}_{off},\widetilde{\overline{t}}_{off})]_{+}))
\end{aligned}$}
\end{equation}
Here $\widetilde{\overline{i}}_{off}$ and $\widetilde{\overline{t}}_{off}$ are the corresponding image and text of $\overline{t}_{off}$ and $\overline{i}_{off}$, respectively. Because $\overline{t}_{off}$ and $\overline{i}_{off}$ are offline hard negatives of corresponding $i$ and $t$, both $(\overline{i}_{off}, \overline{t}_{off})$ and $(\widetilde{\overline{i}}_{off}, \widetilde{\overline{t}}_{off})$ can be also regarded as offline hard negative pairs (we re-sample $\overline{i}_{off}$ and $\overline{t}_{off}$ if they occationally correspond to each other). The samples of each pair are non-corresponding but share very similar semantic meanings to each other, and also to $i$ and $t$. This two new terms guide the model to distinguish between positive and negative pairs without common elements. In Section~\ref{sec:exp}, we prove the effectiveness of deriving the new terms based on $\overline{i}_{off}$, $\overline{t}_{off}$ instead of $\overline{i}_{on}$, $\overline{t}_{on}$. The complete offline loss terms based on anchor $i$ and $t$ contain 4 and 5 elements. Following \cite{chen2017beyond}, we define it as an offline quintuplet loss.

\subsection{Adaptive and Hierarchical Penalization} \label{sec:adap}

In Section~\ref{sec:offline}, we introduce offline hard negatives which cooperate with online hard negatives to train the model as Equation~\ref{equ:onoff}. During the training process, it is natural that we should give different penalty weights to negative pairs with different ``hardness'' degrees. For example, if the similarity score between a positive pair and a hard negative pair is close, both pairs should obtain higher penalty weight which guides the model to distinguish between them better. However, when we derive each loss term with respect to its contained pairs' similarity scores, the gradients are always constant. This indicates that when the margin condition is not satisfied, the penalty weight is consistent regardless of the closeness degree between the positive and negative pairs.

One simple solution is modifying each loss term to a form of square so that the penalty weight is related to the score difference between the positive and negative pairs. However, we find that the improvement is limited as there are no hierarchical knowledge provided by the loss function. Ideally, we expect that the positive pairs to obtain higher scores than offline hard negative pairs, and that the offline hard negative pairs obtain higher scores than online hard negative pairs. To this end, we feed the information of offline hard negatives into the online loss term. The final loss function for the second-round training is as follows:

\begin{equation} \label{equ:final}
\scalebox{0.633}{
$\begin{aligned}
\mathcal{L}= \sum_{(i,t) \in P}(((\beta - \frac{S(i,\overline{t}_{off})-S(i,\overline{t}_{on})}{\alpha})[\gamma_{1} - S(i,t)+S(i,\overline{t}_{on})]_{+} +  [\gamma_{2} - S(i,t)+S(i,\overline{t}_{off})]_{+} + [\gamma_{2} - S(i,t)+S(\overline{i}_{off},\overline{t}_{off})]_{+})  \\ + 
((\beta - \frac{S(\overline{i}_{off},t)-S(\overline{i}_{on},t)}{\alpha})[\gamma_{1} - S(i,t)+S(\overline{i}_{on},t)]_{+} +  [\gamma_{2} - S(i,t)+S(\overline{i}_{off},t)]_{+} + [\gamma_{2} - S(i,t)+S(\widetilde{\overline{i}}_{off},\widetilde{\overline{t}}_{off})]_{+}))
\end{aligned}$}
\end{equation}
Here $\alpha$ and $\beta$ are hyper-parameters. In Section~\ref{sec:exp}, we present that they can be set to consistent values for different models on different datasets. 

To better understand how the proposed loss function works, we focus on the first part (line) of Equation~\ref{equ:final} which is symmetrical to the second part, and compute its gradient with respect to $S(i,t)$, $S(i,\overline{t}_{off})$ and $S(i,\overline{t}_{on})$ as follows:

\begin{equation} 
\scalebox{0.70}{
$\begin{aligned}
\frac{\partial{\mathcal{L}}}{\partial{S(i,t)}} &= (\frac{S(i,\overline{t}_{off})-S(i,\overline{t}_{on})}{\alpha} - \beta)\mathbb{I}(\gamma_{1} - S(i,t)+S(i,\overline{t}_{on})>0) - \mathbb{I}(\gamma_{2} - S(i,t)+S(i,\overline{t}_{off})>0) \\&- \mathbb{I}(\gamma_{2} - S(i,t)+S(\overline{i}_{off},\overline{t}_{off})>0),\\
\frac{\partial{\mathcal{L}}}{\partial{S(i,\overline{t}_{off})}} &= (\frac{S(i,t)-S(i,\overline{t}_{on})}{\alpha} - \frac{\gamma_{1}}{\alpha}))\mathbb{I}(\gamma_{1} - S(i,t)+S(i,\overline{t}_{on})>0) +\mathbb{I}(\gamma_{2} - S(i,t)+S(i,\overline{t}_{off})>0), \\
\frac{\partial{\mathcal{L}}}{\partial{S(i,\overline{t}_{on})}} &= (\frac{2 S(i,\overline{t}_{on}) - S(i,t) - S(i,\overline{t}_{off})}{\alpha} + \beta + \frac{\gamma_{1}}{\alpha})\mathbb{I}(\gamma_{1} - S(i,t)+S(i,\overline{t}_{on})>0)
\end{aligned}$}
\end{equation}
Here $\mathbb{I}(A)$ is the indicator function: $\mathbb{I}(A)$ = 1 if A is true, and 0 otherwise. 

When the margin conditions are not satisfied, the gradient of $\mathcal{L}$ with respect to $S(i,\overline{t}_{on})$ becomes larger when $S(i,\overline{t}_{on})$ is close to the average of $S(i,\overline{t}_{off})$ and $S(i,t)$, which indicates a larger penalty to make $S(i,\overline{t}_{on})$ lower. For the gradient of $\mathcal{L}$ with respect to $S(i,t)$, the second and third terms indicate a negative constant which pushes $S(i,t)$ to be higher than $S(i,\overline{t}_{off})$. In addition, the first term indicates an additional adaptive penalty for $S(i,t)$ to be far away from $S(i,\overline{t}_{on})$. When $S(i,\overline{t}_{on})$ is remarkably lower than $S(i,\overline{t}_{off})$, the penalty drops since $S(i,\overline{t}_{on})$ is sufficiently lower. As for the gradient of $\mathcal{L}$ with respect to $S(i,\overline{t}_{off})$, it is subtle as the second term indicates a positive constant that penalizes $S(i,\overline{t}_{off})$ to be lower than $S(i,t)$. However, this penalty could be neutralized when $S(i,t)$ and $S(i,\overline{t}_{on})$ are close to each other. In this situation, it prevents the penalty from incorrectly making $S(i,\overline{t}_{off})$ lower than $S(i,\overline{t}_{on})$. 

Overall, the proposed loss function applies adaptive and hierarchical penalties to the positive, offline hard negative and online hard negative pairs based on the differences among their predicted scores. Essentially, the pairs that are close to the boundary of its pair class obtain larger penalty weights, the inter-class score gaps can thus be enlarged among these three kinds of pairs. In Section~\ref{sec:exp}, we demonstrate its strong effectiveness to improve the model's performance.

\section{Experiments} \label{sec:exp}

Extensive experiments are performed to evaluate the proposed training approach. The performance of retrieval is evaluated by the standard recall at $K$ (R@K). It is defined as the fraction
of queries for which the correct item belongs to the top-$K$ retrieval items. We first present the datasets, experiment settings and implementation details. We then compare and analyze the performance of the proposed approach with others quantitatively and qualitatively.

\subsection{Dataset and Experiment Settings}
We evaluate our model on two well-known datasets, MS-COCO and Flickr30K. The original MS-COCO dataset \cite{lin2014microsoft} contains 82,783 training and 40,504 validation images. Each image is annotated with five descriptions. Following the splits of \cite{liu2019focus,lee2018stacked,li2019visual}, we divide the dataset into 113,283 training images, 5,000 validation images and 5,000 test images. Following \cite{faghri2017vse++,lee2018stacked,li2019visual}, we report the results by averaging over 5 folds of 1K test images or testing on the full 5K test images. Flickr30k \cite{young2014image} consists of 31K images collected from the Flickr website. Each image also corresponds to five human-annotated sentences. Following the split of \cite{liu2019focus,lee2018stacked,li2019visual}, we randomly select 1,000 images for validation and 1,000 images for testing and use other images to train the model. 

To evaluate the effectiveness and generality of the proposed approach, we apply it to the following current state-of-the-art image-text matching models:

\begin{itemize}
    \item \textbf{SCAN} \cite{lee2018stacked}. The first model that captures image-text correspondence at the level of objects and words. The word and object features are extracted by bi-directional GRU and the combination of Faster R-CNN \cite{ren2015faster} and ResNet-101 \cite{he2016deep}, respectively. Stacked cross attention is fed into the network to discover the full latent alignments using both objects and words as context.

    \item \textbf{BFAN} \cite{liu2019focus}. A novel Bidirectional Focal Attention Network based on SCAN that achieves remarkable improvement. Compared with SCAN, it focuses additionally on eliminating irrelevant fragments from the shared semantics.

    \item \textbf{VSRN} \cite{li2019visual}. The current state-of-the-art image-text matching models without leveraging extra supervision (the model in \cite{ji2019saliency} is trained by extra saliency-annotated data). It generates object representation by region relationship reasoning and global semantic reasoning.
\end{itemize}

All the three models are originally trained by triplet loss with online hard negatives. We replace it with the proposed training approach for comparison.

\subsection{Implementation Details} \label{sec:details}

To perform a fair comparison, for SCAN, BFAN and VSRN, we completely preserve their network structures and model settings (\emph{e.g.} training batch size, feature dimension and other model-related hyper-parameter settings) as described in their original work. We only replace the online triplet loss by the proposed one to train them. For all the situations, the margins for online and offline ranking losses $\gamma_{1}$ and $\gamma_{2}$ are set to 0.2 and 0, the hyper-parameters $\beta$ and $\alpha$ in Equation~\ref{equ:final} are set to 1.5 and 0.3. The top list size $h$ is set to 300 and 60 to sample offline hard negative texts and images (the training texts are 5 times as many training images for both datasets). As mentioned in Section~\ref{sec:offline}, for VSRN, it takes 3,400s/620s to perform global similarity score prediction on MS-COCO/Flickr30K. However, for SCAN and BFAN, they hold complex upper networks which make this step extremely time-consuming. Therefore, we skip the first-round training of SCAN and BFAN. The similarity scores predicted by VSRN are also used as a basis for the second-round training of SCAN and BFAN to sample offline hard negatives. We consider this setting valid because, after the second-round training, the final prediction is still made by SCAN or BFAN without the participating of VSRN, which can be regarded as a teacher model. For the first-round training on MS-COCO/Flickr30K, as \cite{li2019visual}, VSRN is trained by a start learning rate of 0.0002 for 15/10 epochs, and then trained by a lower learning rate of 0.00002 for another 15/10 epochs. For the second-round training on both datasets, SCAN, BFAN and VSRN are trained by a start learning rate of 0.0005, 0.0005 and 0.0002 for 10 epochs, and then trained by a lower learning rate of 0.00005, 0.00005 and 0.00002 for another 5, 5 and 10 epochs, respectively. 

\subsection{Results on MS-COCO and Flickr30K}\label{sec:result}

\begin{table*}[htbp]

  \small\centering
  \caption{\label{tab:result1} Quantitative evaluation results of image-to-text (sentence) retrieval and text-to-image (image) retrieval on MS-COCO 1K/5K test set. The baseline models (first row) are trained by the triplet loss with online hard negatives. ``+ OffTri'', ``+ OffQuin'', ``+ AdapOffQuin'' represent training the model by Equation~\ref{equ:onoffbasic}, \ref{equ:onoff}, \ref{equ:final}, respectively.}
  \begin{threeparttable}
  \scalebox{0.85}{
  \begin{tabular}{|c|c|c|c|c|c|c|c|}

    \cline{1-8}
    &\multicolumn{3}{|c|}{Sentence Retrieval}&\multicolumn{3}{|c|}{Image Retrieval}&\\ 
    Model&R@1&R@5&R@10&R@1&R@5&R@10&rsum \\ \cline{1-8}
    \multicolumn{8}{|c|}{1K Test Images} \\ \cline{1-8}
    SCAN \cite{lee2018stacked} &72.7 &94.8 &98.4 &58.8 &88.4 &94.8 &507.9 \\
    SCAN + OffTri &73.1 &94.8 &98.2 &59.3 &88.3 &94.8 &508.5\\
    SCAN + OffQuin &73.6 &95.0 &98.4 &59.6 &\textbf{88.6} &\textbf{95.0} &510.2 \\
    SCAN + AdapOffQuin &\textbf{74.1} &\textbf{95.2} &\textbf{98.5} &\textbf{59.8} &\textbf{88.6} &\textbf{95.0} &\textbf{511.2}\\\cline{1-8}
    BFAN \cite{liu2019focus} &74.9 &95.2 &98.3 &59.4 &88.4 &94.5 &510.7\\
    BFAN + OffTri &75.8 &95.6 &98.4 &60.1 &88.8 &94.7 &513.4\\
    BFAN + OffQuin &76.3 &95.7 &98.4 &60.5 &89.0 &94.8 &514.7 \\
    BFAN + AdapOffQuin &\textbf{77.3} &\textbf{96.0} &\textbf{98.5} &\textbf{61.2} &\textbf{89.2} &\textbf{95.0} &\textbf{517.2}\\\cline{1-8}
    VSRN \cite{li2019visual} &76.2 &94.8 &98.2 &62.8 &89.7 &95.1 &516.8\\
    VSRN + OffTri &76.8 &95.2 &98.4 &63.1 &89.9 &95.2 &518.6\\
    VSRN + OffQuin &76.9 &95.3 &98.4 &63.3 &90.2 &95.5 &519.7\\
    VSRN + AdapOffQuin &\textbf{77.5} &\textbf{95.5} &\textbf{98.6} &\textbf{63.5} &\textbf{90.5} &\textbf{95.8} &\textbf{521.4}\\\cline{1-8}
    \multicolumn{8}{|c|}{5K Test Images} \\ \cline{1-8}
    SCAN \cite{lee2018stacked} &50.4 &82.2 &90.0 &38.6 &69.3 &\textbf{80.4} &410.9\\
    SCAN + AdapOffQuin &\textbf{51.2} &\textbf{82.5} &\textbf{90.1} &\textbf{39.4} &\textbf{69.7} &\textbf{80.4} &\textbf{413.3} \\\cline{1-8}
    BFAN \cite{liu2019focus} &52.9 &82.8 &90.6 &38.3 &67.8 &79.3 &411.7\\
    BFAN + AdapOffQuin &\textbf{57.3} &\textbf{84.5} &\textbf{91.7} &\textbf{40.1} &\textbf{69.2} &\textbf{80.1} &\textbf{422.9} \\\cline{1-8}
    VSRN \cite{li2019visual} &53.0 &81.1 &89.4 &40.5 &70.6 &81.1 &415.7\\
    VSRN + AdapOffQuin &\textbf{55.1} &\textbf{83.3} &\textbf{90.8} &\textbf{41.1} &\textbf{71.5} &\textbf{82.0} &\textbf{423.8} \\\cline{1-8}

    \end{tabular}}%

  \end{threeparttable}
\end{table*}%

Table~\ref{tab:result1} shows the performance comparison of models trained by different approaches on MS-COCO. We can see that all the three models are significantly improved on all the settings when trained by our proposed training approach. As mentioned in Section~\ref{sec:details}, for all the models, the offline hard negatives in their second-round training are sampled from the prediction of the first-round trained VSRN. It indicates that the proposed training approach is insensitive to the model consistency of the two-round training. When the global similarity score prediction step is intractable for the current model, we can train it by sampling offline hard negatives based on the prediction of another more efficient model. Overall, we achieve the most significant improvement on BFAN. In particular, on the more reliable 5K test set, it outperforms the baseline by 8.3\% and 4.7\% in top-1 sentence retrieval and top-1 image retrieval.

Table~\ref{tab:result2} shows the performance comparison on Flickr30K. It should be noted that Flickr30K is much smaller than MS-COCO as it contains fewer very confusing negative image-text pairs to be served as high-quality offline hard negative pairs. However, significant improvements are still observed for all the models. In Section~\ref{sec:abl}, we show that our proposed training approach has strong robustness for the quality of offline hard negatives.

\begin{table*}[htbp]

  \small\centering
  \caption{\label{tab:result2} Quantitative evaluation results of sentence retrieval and image retrieval on the Flickr30K test set.}
  \begin{threeparttable}
  \scalebox{0.85}{
  \begin{tabular}{|c|c|c|c|c|c|c|c|}

    \cline{1-8}
    &\multicolumn{3}{|c|}{Sentence Retrieval}&\multicolumn{3}{|c|}{Image Retrieval}&\\ 
    Model&R@1&R@5&R@10&R@1&R@5&R@10&rsum \\ \cline{1-8}
    \multicolumn{8}{|c|}{1K Test Images} \\ \cline{1-8}
    SCAN \cite{lee2018stacked} &67.4 &90.3 &\textbf{95.8} &48.6 &77.7 &85.2 &465.0\\
    SCAN + AdapOffQuin &\textbf{70.3} &\textbf{92.0} &95.5 &\textbf{50.0} &\textbf{79.2} &\textbf{86.2} &\textbf{473.2}\\\cline{1-8}
    BFAN \cite{liu2019focus} &68.1 &91.4 &95.9 &50.8 &78.4 &85.8 &470.4\\
    BFAN + AdapOffQuin &\textbf{73.2} &\textbf{94.5} &\textbf{97.0} &\textbf{54.0} &\textbf{80.3} &\textbf{87.7} &\textbf{486.7} \\\cline{1-8}
    VSRN \cite{li2019visual} &71.3 &90.6 &\textbf{96.0} &54.7 &81.8 &88.2 &482.6\\
    VSRN + AdapOffQuin &\textbf{72.8} &\textbf{91.8} &95.8 &\textbf{55.3} &\textbf{82.2} &\textbf{88.4} &\textbf{486.3}\\\cline{1-8}

    \end{tabular}}%

  \end{threeparttable}
\end{table*}%

\begin{figure}[!t]

\centering
\includegraphics[width=0.9\columnwidth]{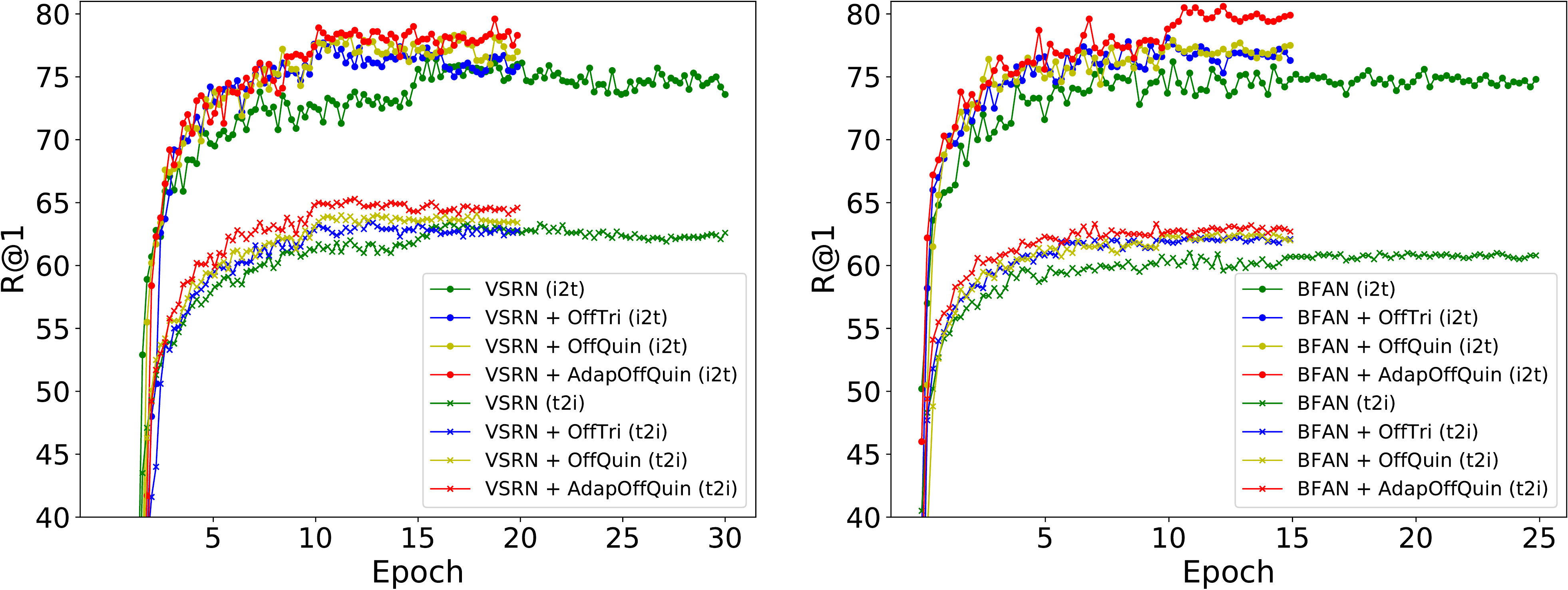}

\caption{Plotting training epoch against R@1 on the MS-COCO validation set for different training approaches applied on VSRN and BFAN. For the proposed approaches, the training curves correspond to the second-round training. ``t2i'' and ``i2t'' represents image retrieval and sentence retrieval, respectively.}
\label{fig:curve}

\end{figure}

We look deeper into different training approaches by examining VSRN and BFAN's training behaviours\footnote{The final BFAN model is an ensemble of two independently trained models BFAN-equal and BFAN-prob \cite{liu2019focus}, here we show the behaviours of BFAN-prob.} on the widely-used MS-COCO 1K validation set \cite{faghri2017vse++,li2019visual,lee2018stacked} (\emph{i.e.} the first fold of the 5K validation set). As shown in Figure~\ref{fig:curve}, both models' performance obtains continuous improvement as we feed different proposed mechanisms into the training process. When the models are trained by Equation~\ref{equ:final}, they converge significantly faster than the baselines as it takes less than 10 epochs for them to outperform the highest R@1 of their baselines.

\subsection{Ablation Study and Visualization}\label{sec:abl}

\begin{table*}[htbp]

  \small\centering
  \caption{\label{tab:result3} Performance of different training approach variants on MS-COCO 1K test set. ``OnlyOffline'' represents the model that is only trained by the offline term. ``Fine-tune'' represents the model that is fine-tuned in the second-round instead of re-trained from scratch. ``OnlineQuin'' indicates that we apply online quintuplet loss instead of offline in Equation~\ref{equ:onoff} (\emph{i.e} replace $S(\overline{i}_{off},\overline{t}_{off}))$), $S(\widetilde{\overline{i}}_{off},\widetilde{\overline{t}}_{off}))$ with $S(\overline{i}_{on},\overline{t}_{on}))$), $S(\widetilde{\overline{i}}_{on},\widetilde{\overline{t}}_{on}))$) to train the model. ``w/o OfflineAdap'' represents that we replace $S(i,\overline{t}_{off})$ and $S(\overline{i}_{off},t)$ by $S(i,t)$ for the new added terms in Equation~\ref{equ:final} to train the model. Performance of selecting different top list size $h$ for offline hard negative text sampling is also studied. The values in parentheses indicate the performance difference between the models trained by the variant and by the proposed approach with the final settings.}
  \begin{threeparttable}
  \scalebox{0.77}{
  \begin{tabular}{|c|c|c|c|c|c|c|}

    \cline{1-7}
    &\multicolumn{3}{|c|}{Sentence Retrieval}&\multicolumn{3}{|c|}{Image Retrieval}\\ 
    Model&R@1&R@5&R@10&R@1&R@5&R@10 \\ \cline{1-7}
    \multicolumn{7}{|c|}{1K Test Images} \\ \cline{1-7}
    
    BFAN (OnlyOffline) &1.1(-76.2) &2.5 (-93.5)&4.9 (-93.6)&0.5 (-60.7)&1.4 (-87.8)&2.6(-92.4)\\
    VSRN (OnlyOffline) &0.7(-76.8) &2.1(-93.4) &3.8(-94.8) &0.4(-63.1) &1.2(-89.3) &2.3(-93.5)\\\cline{1-7}
    BFAN (Fine-tune) &74.3 (-3.0)&94.7 (-1.3)&98.2 (-0.3)&58.7 (-2.5)&88.1 (-1.1)&94.2(-0.8)\\
    VSRN (Fine-tune) &74.5 (-3.0)&94.3 (-1.2)&98.1 (-0.5)&62.0 (-1.5)&89.3(-1.2) &94.8(-1.0)\\\cline{1-7}
    BFAN (OnlineQuin) &75.3 (-2.0)&95.8 (-0.2)&98.5 (+0.0)&59.8 (-1.4)&88.6 (-0.6)&94.6(-0.4)\\
    VSRN (OnlineQuin) &76.4 (-1.1)&94.9 (-0.6)&98.2 (-0.4)&62.8 (-0.7)&89.9(-0.6) &95.2(-0.6)\\\cline{1-7}
    BFAN (w/o OfflineAdap) &76.6 (-0.7)&95.8(-0.2) &98.4 (-0.1)&60.8 (-0.4)&89.1 (-0.1)&94.8(-0.2)\\
    VSRN (w/o OfflineAdap) &77.1 (-0.4)&95.4(-0.1) &98.4 (-0.2)&63.4 (-0.1)&90.2 (-0.3)&95.5(-0.3)\\\cline{1-7}
    VSRN (h = 200) &77.1 (-0.4)&95.3(-0.2) &98.4 (-0.2)&63.3 (-0.2)&90.4 (-0.1)&95.6(-0.2)\\
    VSRN (h = 500) &77.4 (-0.1)&95.6(+0.1) &98.6 (+0.0)&63.5 (+0.0)&90.4 (-0.1)&95.7(-0.1)\\
    VSRN (h = 1000) &77.3 (-0.2)&95.4(-0.1) &98.6 (+0.0)&63.3 (-0.2)&90.3 (-0.2)&95.6(-0.2)\\\cline{1-7}

    \end{tabular}}%

  \end{threeparttable}
\end{table*}%

\begin{figure}[!t]

\centering
\includegraphics[width=0.98\columnwidth]{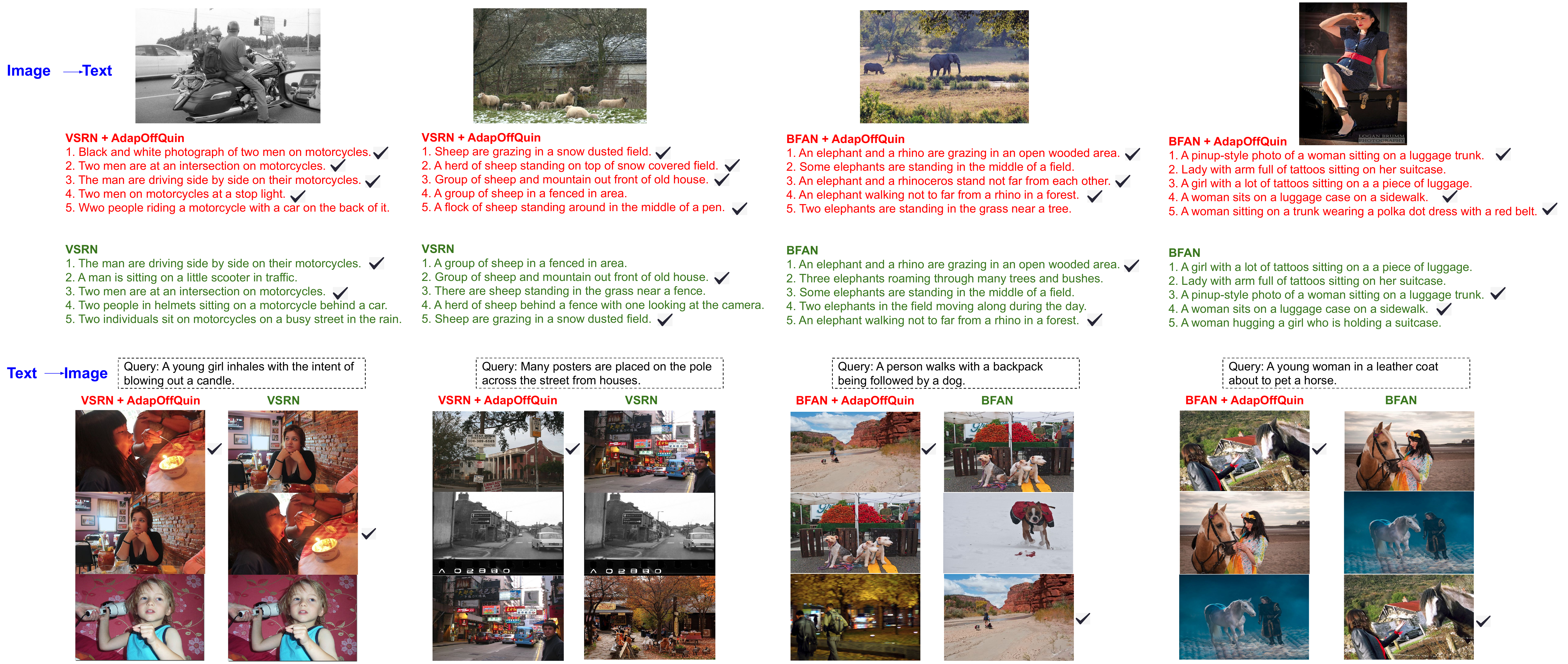}

\caption{Qualitative image retrieval and sentence retrieval comparison between the baseline training approach and ours on the MS-COCO test set.}
\label{fig:example}

\end{figure}

First, we validate whether the offline hard negatives can completely replace online hard negatives to train the model. Specifically, we remove the online loss term in Equation~\ref{equ:onoff} to train VSRN and BFAN. As shown in Table~\ref{tab:result3}, the training process fails as it is too difficult for the model to directly learn to distinguish between the positive pairs and these extremely confusing negative pairs. Also, we demonstrate the usefulness of re-training the model from scratch in the second round. As shown in Table~\ref{tab:result3}, when we apply Equation~\ref{equ:final} to fine-tune the model that has already been trained by the online triplet loss and get trapped in a local optimum, it cannot obtain additional improvement. In Equation~\ref{equ:onoff}, we create two new terms based on offline negatives. Indeed, we can instead apply them based on online negatives. However, the performance of ``OnlineQuin'' models are remarkably worse than the models train by Equation~\ref{equ:onoff}, this supports our claim of the second problem in Section~\ref{sec:intro}. On the other hand, in Equation~\ref{equ:final}, we feed the offline hard negative information into the online term for hierarchical penalization. To validate its effectiveness, we replace $S(i,\overline{t}_{off})$ and $S(\overline{i}_{off},t)$ by $S(i,t)$ for the new added terms in Equation~\ref{equ:final} to break this hierarchical relation. $\alpha$ and $\beta$ are re-adjusted to achieve the best performance on the validation set. The performance drops to the same level of using Equation~\ref{equ:onoff} to train the models, indicating the effectiveness. In the end, for VSRN, we present the model's performance when selecting different top list size $h$ for offline hard negative text sampling (we always keep it 5 times larger than the top list size for offline hard negative image sampling). We can find that even when $h$ is set to $1000$ which indicates significant drops of ``hardness'' degree of offline hard negatives, the model still achieves great performance. This is consistent with the excellent performance on Flickr30K and proves the robustness of our training approach on smaller datasets when very confusing hard negative pairs are limited.

Figure~\ref{fig:example} shows the qualitative comparison between the models trained by different approaches on MS-COCO. For sentence retrieval, given an image query, we show the top-$5$ retrieved sentences. For image retrieval, given a sentence query, we show the top-$3$ retrieved images. The correct retrieval items for each query are ticked off. Overall, our training approach guides the model to better find and attend to the detailed non-correspondences of negative image-text pairs such as  ``snow covered field'', ``rhiho'', ``blowing out a candle'' and ``poster''.

\section{Conclusion}

We present a novel training approach for image-text matching. It starts by mining  ``harder'' negatives offline from the whole training set. Based on the mined offline hard negatives, an effective quintuplet loss is proposed to complement the online triplet loss to better distinguish positive and negative pairs. Furthermore, we take the distance relations among positive, offline hard negative and online hard negative pairs into consideration and effectively achieve adaptive penalization for different pairs. Extensive experiments demonstrate the effectiveness and generality of the proposed approach.

\section{Acknowledgment}
This work is supported in part by NSF awards IIS-1704337, IIS-1722847, and IIS-1813709, as well as our corporate sponsors.

\section{Supplementary Material}
In this document, we provide additional materials to supplement our paper ``Adaptive Offline Quintuplet Loss for Image-Text Matching''. In the first section, we perform additional ablation studies to verify the robustness and efficiency of the proposed training approach. In the second section, we show additional
qualitative examples to compare the performance of the models trained by different approaches.  

\subsection{Ablation Study}
In the section, we perform extra experiments to demonstrate the robustness and efficiency of our proposed training approach. All the experiments are performed based on VSRN (the last one is based on both VSRN and BFAN) on a single GeForce GTX 1080 Ti GPU. 

First, one may ask whether simply increasing the training mini-batch size can replace our proposed training approach since it can also increase the ``hardness'' of the negative samples. To verify this, we increase the training batch size of VSRN to 192 -- the maximum batch size that can be allocated by a single GeForce GTX 1080 Ti GPU. As shown in Table~\ref{tab:result6}, simply improving the batch size cannot lead to better performance, demonstrating the effectiveness and validity of the proposed training approach.   

\begin{table*}[htbp]
  \small\centering
  \caption{\label{tab:result6} Performance of VSRN with the mini-batch size set to 192.}
  \begin{threeparttable}
  \scalebox{0.9}{
  \begin{tabular}{|c|c|c|c|c|c|c|}

    \cline{1-7}
    &\multicolumn{3}{|c|}{Sentence Retrieval}&\multicolumn{3}{|c|}{Image Retrieval}\\ 
    Model&R@1&R@5&R@10&R@1&R@5&R@10\\ \cline{1-7}
    \multicolumn{7}{|c|}{1K Test Images} \\ \cline{1-7}
    VSRN &75.7&95.0&98.3&62.4&89.7&95.2 \\  \cline{1-7}
    
    \end{tabular}}%
  \end{threeparttable}
\end{table*}%

In Section~\ref{sec:adap}, we feed the information of offline hard negatives into the online loss term and present a new loss function (\emph{i.e.} Equation 5 of the main paper) for the second-round training. Hyper-parameters $\alpha$ and $\beta$ are used to adjust the degree of adaptive penalization. Table~\ref{tab:result4} shows the performance of training VSRN with different $\alpha$ and $\beta$ on the MSCOCO 1K test set. Overall, the performance difference is little when using different $\alpha$ and $\beta$ of specific ranges to train the model. The adaptive penalization is not very sensitive to the selection of $ \alpha $ and $ \beta $ and constantly makes a positive effect on the training process, indicating the robustness of the proposed approach in Section~\ref{sec:adap}.

\begin{table*}[htbp]
  \small\centering
  \caption{\label{tab:result4} Performance of selecting different $\alpha$ and $\beta$ for adaptive penalization.}
  \begin{threeparttable}
  \scalebox{0.9}{
  \begin{tabular}{|c|c|c|c|c|c|c|}

    \cline{1-7}
    &\multicolumn{3}{|c|}{Sentence Retrieval}&\multicolumn{3}{|c|}{Image Retrieval}\\ 
    Model&R@1&R@5&R@10&R@1&R@5&R@10\\ \cline{1-7}
    \multicolumn{7}{|c|}{1K Test Images} \\ \cline{1-7}
    VSRN ($\beta=1.5$, $\alpha=0.2$)  &77.1&95.3&98.7&63.8&90.6&95.8 \\ 
    VSRN ($\beta=1.5$, $\alpha=0.3$) &77.5&95.5&98.6&63.5&90.5&95.8 \\ 
    VSRN ($\beta=1.5$, $\alpha=0.5$) &77.5&95.5&98.5&63.3&90.3&95.6 \\ 
    VSRN ($\beta=1.0$, $\alpha=0.3$) &77.0&95.2&98.5&63.2&90.1&95.5 \\ 
    VSRN ($\beta=2.0$, $\alpha=0.3$) &77.4&95.4&98.7&63.3&90.3&95.7 \\ \cline{1-7}
    
    \end{tabular}}%
  \end{threeparttable}
\end{table*}%

In addition, to evaluate the training efficiency of the proposed training approach (the model's inference efficiency is unrelated to the training approaches), we compare the per-batch training time of VSRN with different training approaches on a fixed mini-batch size of 128. As shown in Table~\ref{tab:result5}, when we employ our proposed approach, the training speed drops since the model needs to additionally compute the similarity score between the anchor and its sampled offline hard negatives. Overall, the per-batch training speed of VSRN with our proposed training approach is about 1.5 times slower than the per-batch training speed of VSRN with the basic online triplet loss. Considering that VSRN needs to be trained for 30 epochs (first round) by the online triplet loss and 20 epochs (second round) by the proposed training approach, the total training time of the second round is acceptable.

\begin{table*}[htbp]
  \small\centering
  \caption{\label{tab:result5} Training efficiency comparison among different training approaches.}
  \begin{threeparttable}
  \scalebox{0.9}{
  \begin{tabular}{|c|c|}

    \cline{1-2}
     
    Model&Per-batch Training Time (Second)\\ \cline{1-2}
    \multicolumn{2}{|c|}{1K Test Images} \\ \cline{1-2}
    VSRN  &1.096\\ 
    VSRN + OffTri&1.489\\ 
    VSRN + OffQuin&1.557\\ 
    VSRN + AdapOffQuin&1.604\\ \cline{1-2}
    
    \end{tabular}}%
  \end{threeparttable}
\end{table*}%

\begin{figure}

\centering
\includegraphics[width=0.9\columnwidth]{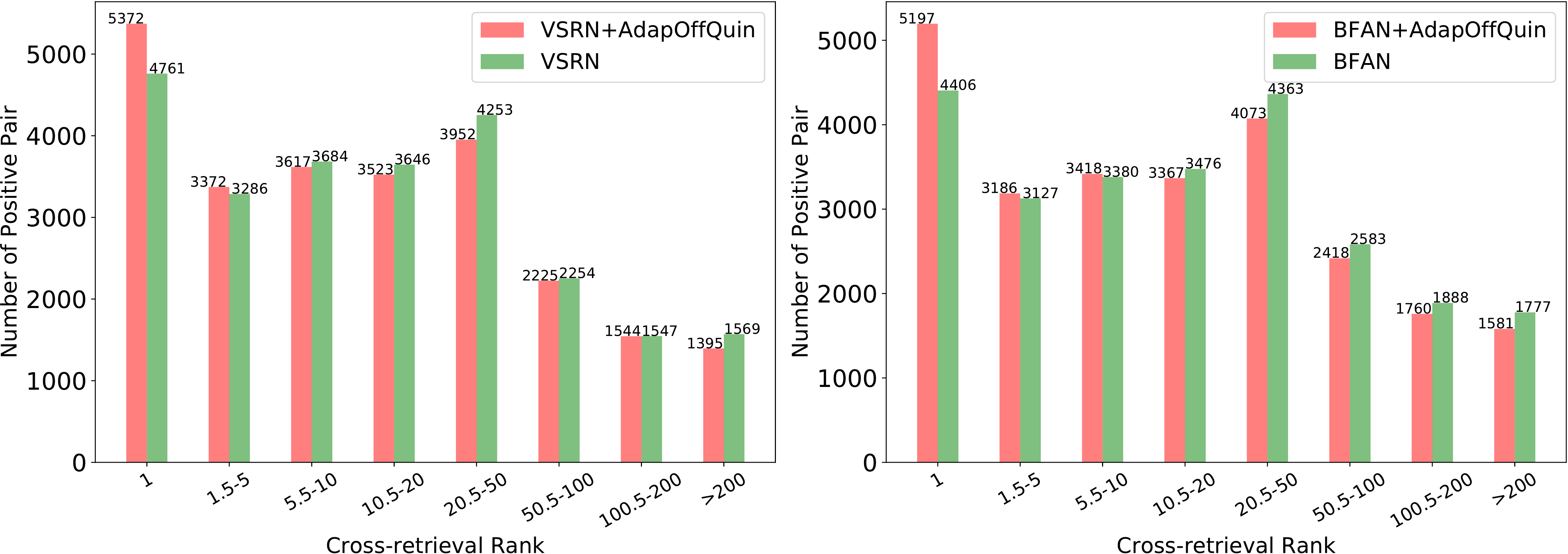}
\caption{Comparison of positive pairs' cross-retrieval rank frequency on the MS-COCO test set for different training approaches applied on VSRN and BFAN.}
\label{fig:pairsta}

\end{figure}

In the end, for our cross-modality retrieval task, a corresponding positive image-text pair may perform well on one modality but poorly on the other (\emph{e.g.} ranks top against the negative pairs that share the same image, but obtains low rank against the negative pairs that share the same text). We prove that our training approach does not exacerbate this unbalance. On the full MS-COCO 5K test set that contains 5,000 images, 25,000 texts, and 25,000 positive image-text pairs, for each pair, the trained models predict its rank against the 4,999 negative pairs that share the same text and 24,995 negative pairs that share the same image as $r_{i}$ and $r_{t}$. For fair weighting between $r_{i}$ and $r_{i}$ with different negative pair numbers, the cross-retrieval rank of each positive pair is defined as: $\max(5r_{i}-4,  r_{t})$. It records the lower rank of the positive pair against the two kinds of negative pairs. Figure~\ref{fig:pairsta} shows the 25,000 positive pairs' cross-retrieval rank frequency distribution of different rank intervals. It can be seen that for both VSRN and BFAN, the number of positive pairs with the cross-retrieval rank of $1$ (\emph{i.e.} the positive pair's score is higher than the scores of all the 4,999 text-shared and 24,995 image-shared negative pairs) increases significantly when the proposed approach is applied. Meanwhile, the number drops for the pairs with cross-retrieval rank larger than 200, indicating a comprehensive improvement for the overall ranking of positive pairs in the test set.

It should be noticed that we do not specially handle the false negative problem in the dataset -- for some common scenes that occur many times (e.g., "surfing man"), there are offline negatives that should be considered positive. We instead implicitly avoid frequently sampling them by setting the top list size ``h'' to be not too small. For most anchors, there are no more than five false negatives in the dataset. The final setting in Section~\ref{sec:details} can safely maintain a very low rate of false negatives and prevent them from having a bad effect. Moreover, we also try to sample the negatives from a normal distribution instead of a uniform distribution to reduce the probability of sampling the most top ones that could be false negatives. We found that it does not lead to further improvement when ``h'' has already been set to a suitable value. The performance difference between sampling from a normal distribution or a uniform distribution is little.

\subsection{Additional Qualitative Results}

\begin{figure}[!t]

\centering
\includegraphics[width=1\columnwidth]{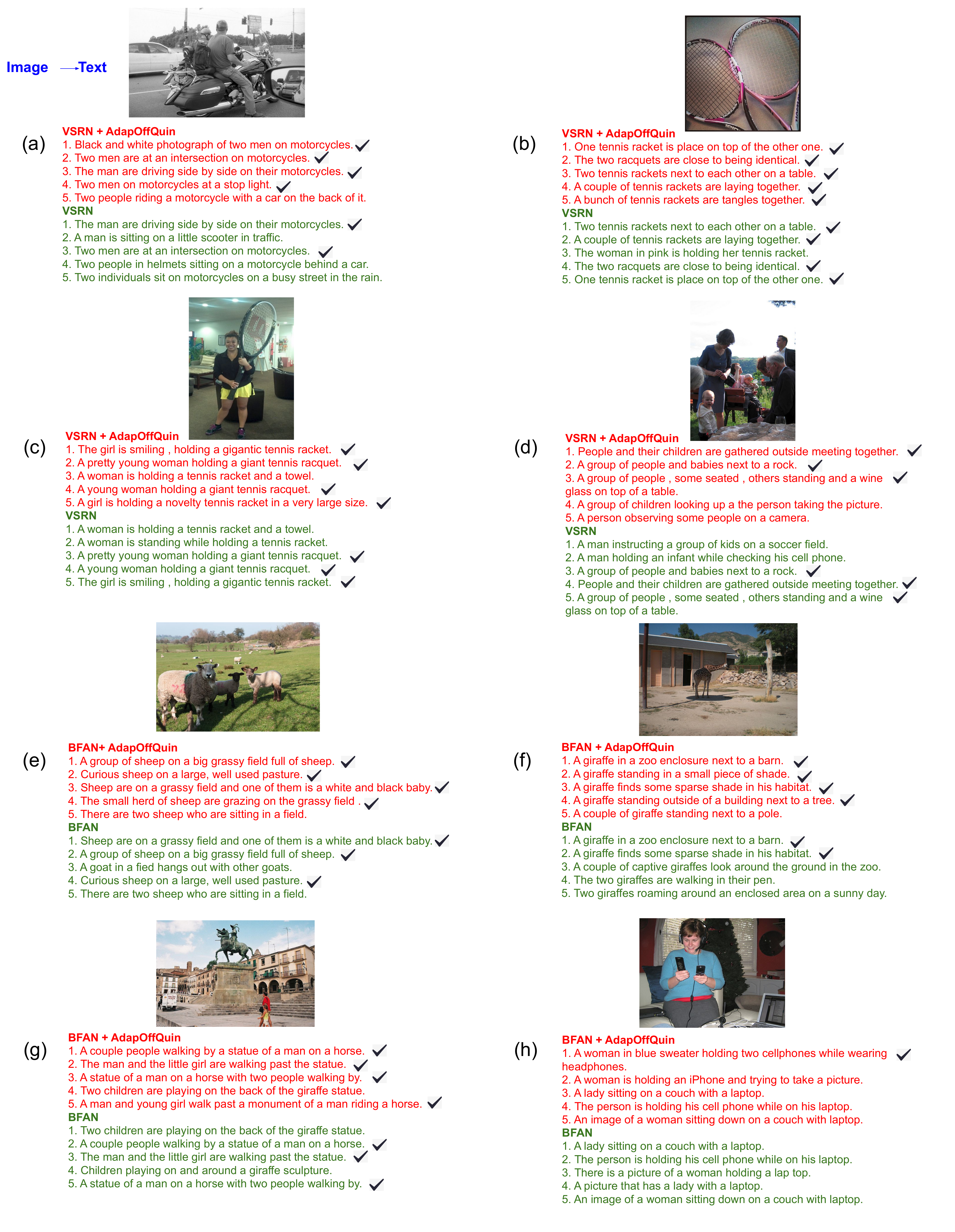}

\caption{Qualitative sentence retrieval comparison between the baseline training approach and ours on the MS-COCO test set.}
\label{fig:sup1}

\end{figure}

\begin{figure}[!t]

\centering
\includegraphics[width=1\columnwidth]{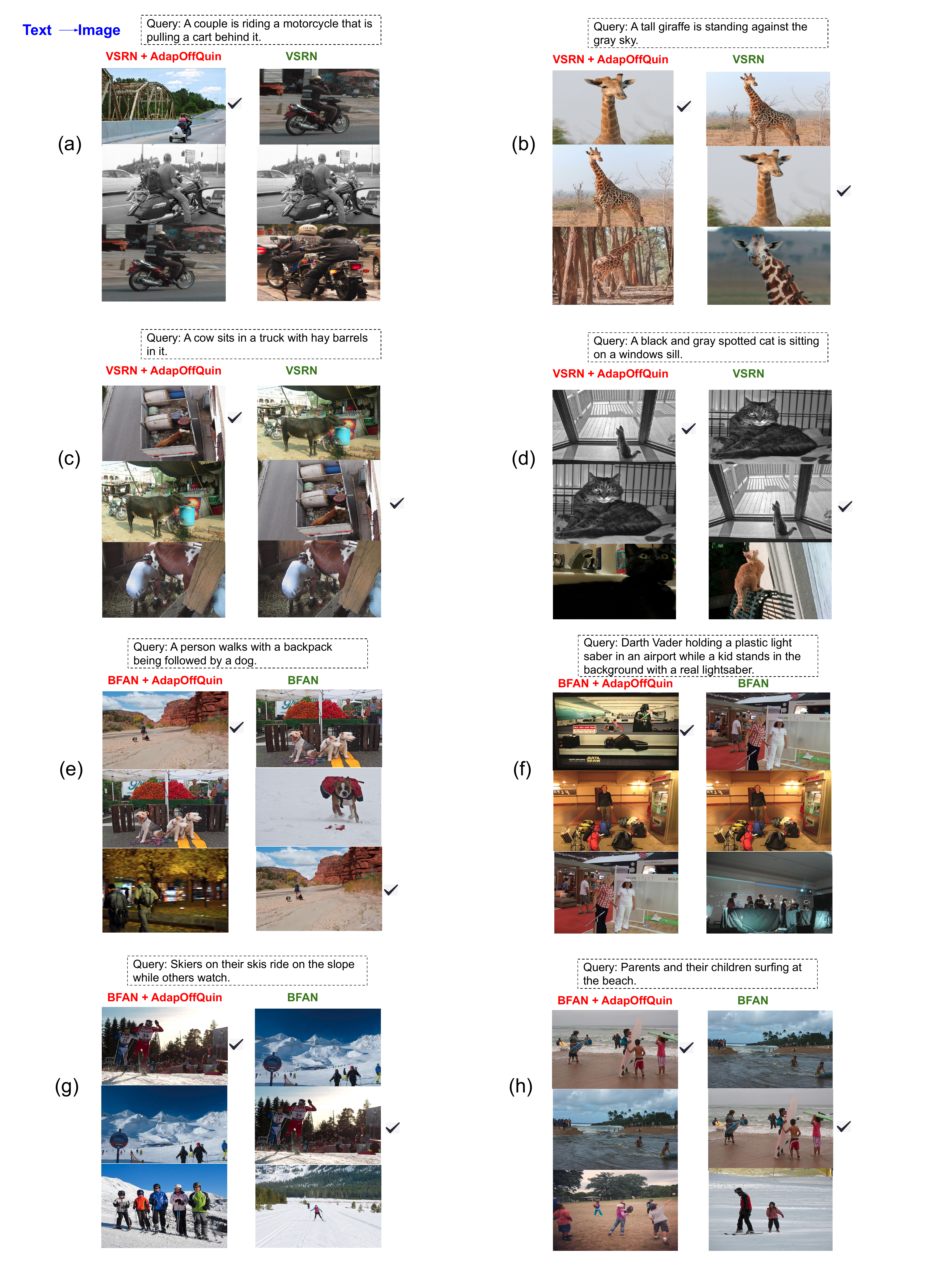}

\caption{Qualitative image retrieval comparison between the baseline training approach and ours on the MS-COCO test set.}
\label{fig:sup2}
\end{figure}

In this section, we provide a great number of image retrieval and sentence retrieval examples to compare the models trained by the baseline approach and the proposed one. Qualitative sentence and image retrieval results are shown as in Figure~\ref{fig:sup1} and Figure~\ref{fig:sup2}, respectively. From Figure~\ref{fig:sup1}, the models trained by the proposed approach achieve better performance to differentiate between the corresponding sentences and the confusing non-corresponding sentences of an image query. In particular, they perform better to find the detailed non-correspondences of non-corresponding image-text pairs from the object number (\emph{e.g.} ``the man'' in (a)), object attribute (\emph{e.g.} ``goat'' in (e)) and object relation (\emph{e.g.} ``holding a laptop'' in (h)). As for image retrieval, as shown in Figure~\ref{fig:sup2}, they can successfully identify the images that miss the corresponding information (\emph{e.g.} ``in a truck'' in (c), ``real lightsaber'' in (f),  ``skis ride'' in (g)) or contain the false information (\emph{e.g.} ``gray sky'' in (b)).

\clearpage
%
%
\bibliographystyle{splncs04}
\bibliography{egbib}
\end{document}